\documentclass[letterpaper, 10 pt, conference]{ieeeconf} %

\IEEEoverridecommandlockouts %

\newcommand{\projectsite}{https://sttawm.github.io/rephrase-before-you-act}
\newcommand{\artifactsrepo}{https://github.com/sttawm/vla-rephrasing-artifacts}

\usepackage{cite}
\usepackage{amsmath,amssymb,amsfonts}
\usepackage{graphicx}
\usepackage{textcomp}
\usepackage{xcolor}
\usepackage{booktabs}
\usepackage{multirow}
\usepackage{array}
\usepackage{url}
\usepackage{algorithm}
\usepackage{algpseudocode}
\usepackage[caption=false,font=footnotesize]{subfig}
\newcommand{\figcaptionfont}{\small}
\makeatletter
\let\IEEE@makecaption\@makecaption
\long\def\@makecaption#1#2{\begingroup
 \ifx\@captype\@IEEEtablestring\else\let\footnotesize\figcaptionfont\fi
 \IEEE@makecaption{#1}{#2}\endgroup}
\makeatother
\definecolor{todocolor}{RGB}{190,0,160}

\usepackage{graphics} %
\usepackage{epsfig} %
\usepackage{amsmath} %
\usepackage{amssymb} %

\usepackage[utf8]{inputenc} %
\usepackage[T1]{fontenc} %

\usepackage[dvipsnames]{xcolor}
\usepackage{booktabs} %
\usepackage{amsfonts} %
\usepackage{nicefrac} %
\usepackage{microtype} %

\usepackage{booktabs} %
\usepackage{multirow}
\usepackage{pifont}
\usepackage{setspace}
\usepackage{comment}
\usepackage{algorithm}

\usepackage{verbatim}
\usepackage{caption}
\usepackage{wrapfig}
\usepackage{svg}
\usepackage{mathabx}
\usepackage{subcaption}
\makeatletter
\let\phr@newlength\newlength
\def\newlength#1{\ifx#1\labelindent\else\phr@newlength#1\fi}
\usepackage{enumitem}
\let\newlength\phr@newlength
\makeatother
\usepackage{color,soul}
\usepackage{xfakebold}

\setul{0.5ex}{0.3ex}
\definecolor{grey}{rgb}{.80,.80,0.80}
\setulcolor{grey}

\usepackage{hyperref}
\hypersetup{
 colorlinks=true,
 linkcolor=orange,
 filecolor=magenta,
 urlcolor=orange,
 citecolor=orange,
}

\usepackage[T1]{fontenc}
\usepackage{graphicx}
\usepackage{xcolor}
\usepackage{listings}
\usepackage{url}
\usepackage{capt-of}
\usepackage{booktabs}
\usepackage{balance}

\definecolor{hi}{RGB}{212,241,212}
\definecolor{lo}{RGB}{250,214,214}
\definecolor{hipct}{RGB}{22,122,39}
\definecolor{lopct}{RGB}{178,32,32}
\definecolor{mid}{RGB}{236,236,236}
\definecolor{midpct}{RGB}{90,90,90}

\title{\LARGE \bf
Rephrase Before You Act: Characterizing and Mitigating Language Sensitivity in Vision-Language-Action Models
}

\author{%
Mikey Watts$^{1}$ and Yuchen Cui$^{2}$%
\thanks{$^{1}$Mikey Watts is an independent researcher in Los Angeles
{\tt\small mikeyMWwatts@gmail.com}}%
\thanks{$^{2}$Yuchen Cui is with the University of California, Los Angeles
{\tt\small yuchencui@cs.ucla.edu}}%
}

\graphicspath{{./}{./figures/}}

\begin{document}

\maketitle

\thispagestyle{empty}
\pagestyle{empty}

\begin{abstract}
Vision-language-action models (VLAs) are strikingly sensitive to
instruction phrasing and do not inherit the language robustness of the
vision-language models they are built on. A one-word edit can move success by tens of points: $\pi_{0.5}$ turns on a LIBERO stove 100\% of the time for ``switch on the stove'' and 2\% for ``switch on the hot plate'', and a $\pi_0$ checkpoint finetuned with rephrase augmentation still shows swings of up to 61 points. We characterize this sensitivity
with statistically tested single-edit swings and an oracle phrase search,
which shows that phrasing alone nearly closes the 21-point gap between
in-distribution and out-of-distribution tasks. We then reduce it without
modifying the policy. Because the sensitivity is systematic, it can be expressed as explicit rules: we score many phrasings of a few training tasks, have a large language model distill the evidence into ten to twenty rephrasing rules, and at deployment rewrite each incoming instruction once under these rules. The rules improve the frozen $\pi_0$ by 16 to 27\% relative
on twelve held-out tasks across adversarial, VLM-generated, and
human-generated phrasings, with gains concentrated on out-of-distribution tasks. The pipeline replicates on $\pi_{0.5}$ and LIBERO, lifting
in-finetune success from 93.6\% to 97.8\%. The method requires no
retraining and no per-step verification, and applies zero-shot to unseen
tasks and instructions.
Project website:
\url{\projectsite}
\end{abstract}

\section{Introduction}

General-purpose robots that average end users can direct in their own words remain a long-standing goal of the field, and such users cannot be expected to phrase a request the way the robot was trained. Generalist vision-language models (VLMs) have progressed rapidly, and vision-language-action models (VLAs) extend them to robot control in the hope of inheriting their capabilities, including robustness to how an instruction is phrased. This hope is often not met. Telling $\pi_{0.5}$ to \textit{``switch on the stove''} succeeds 100\% of the time, whereas \textit{``switch on the hot plate''} succeeds 2\% of the time, and telling $\pi_0$ that the \textit{``purple eggplant goes on the sponge''} succeeds 69\% of the time on a SIMPLER task, whereas dropping the color adjective lowers the success rate to 8\%.

A robot that executes a task only under particular wording is not a robot that non-experts can instruct. The standard remedy is training-time augmentation, which relabels demonstrations with additional phrasings~\cite{kwok2026scaling, xiao2023dial}. The $\pi_0$ checkpoint we study was finetuned with rephrase augmentation~\cite{intact}, which is worth a 20\% relative improvement on naturally phrased requests in our measurement, yet it still exhibits the swings above. This residual sensitivity is the problem we address, and existing deployment-side methods do not close it. Test-time verification~\cite{kwok2025robomonkey, kwok2026scaling} selects among many phrase-action candidates at every step, which multiplies inference cost without improving over the unmodified policy in our setting. Instruction optimization~\cite{bhatia2026sarl, shodiev2026grounder, myers2024palo} needs online interaction with each task, so nothing transfers to an unseen one. No existing method leaves the policy's weights untouched, acts once per episode, and transfers zero-shot to tasks it has never seen.

\begin{figure*}[!t]
 \vspace*{4pt}%
 \centering
 \includegraphics[width=\textwidth, height=0.5\textheight, keepaspectratio]{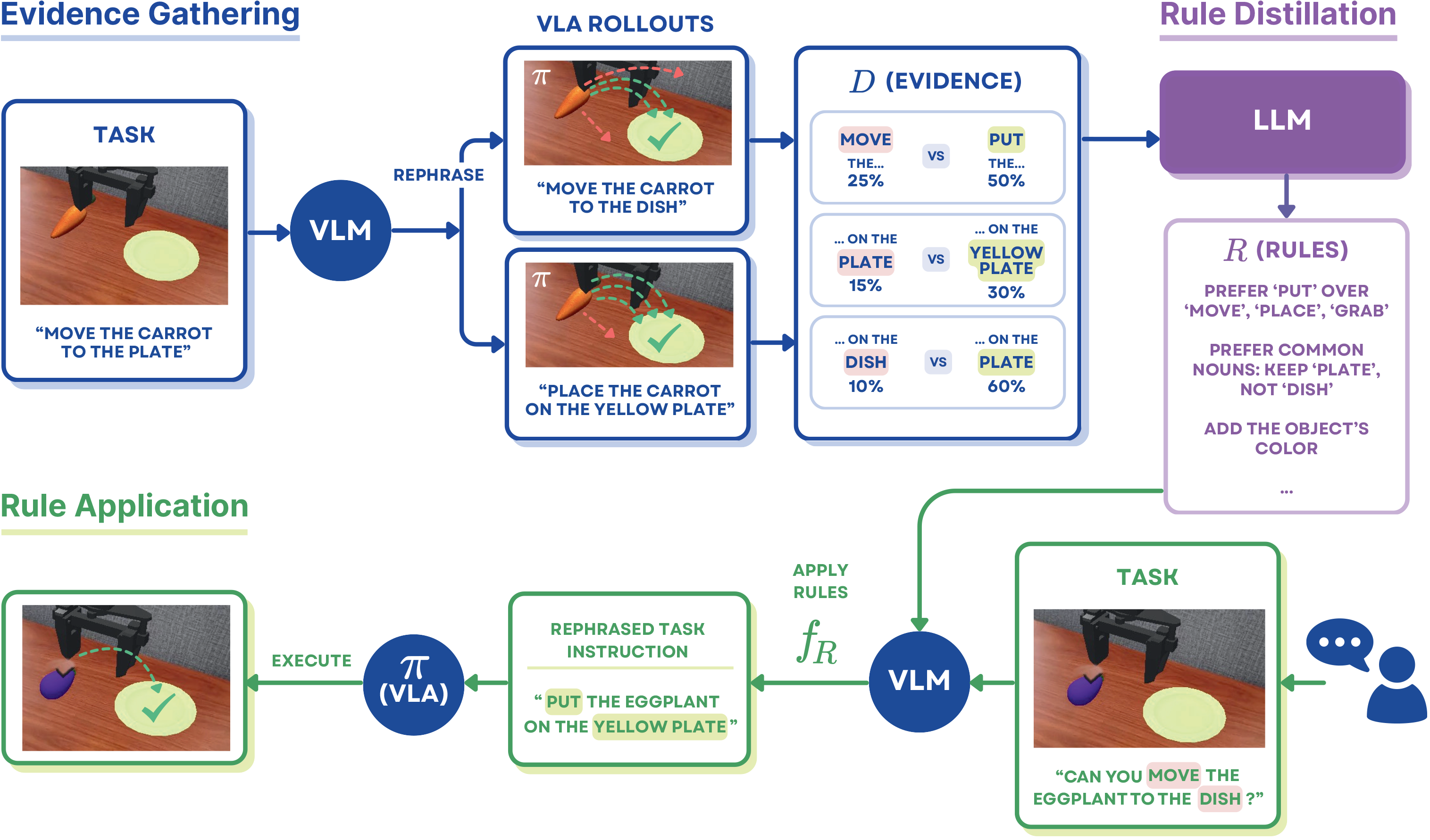}
 \caption{\small Method overview. \emph{Evidence gathering}: a VLM rephrases each training task, and the frozen VLA's rollout success on each phrasing forms the evidence $\mathcal{D}$. \emph{Rule distillation}: an LLM distills $\mathcal{D}$, once and offline, into 10--20 explicit phrasing rules $R$. \emph{Rule application}: at deployment a VLM conditioned on $R$ rewrites each incoming instruction once ($f_R$) before passing it to the unchanged VLA. The phrasings, scores, and rules shown are illustrative only.}
 \label{fig:method}
\end{figure*}

Our key observation is that a VLA's language sensitivity is systematic rather than idiosyncratic to each task, and can therefore be captured by a small set of explicit rules. We generate many phrasings of eight training tasks, roll each out in simulation, and have a large language model distill the scored phrasings into ten to twenty natural-language rules, for example \textit{```coke' stays `coke', never `cola', `soda', or bare `can' ''} and \textit{``never swap a common noun for its sibling (keep 'plate', not 'dish')''}. At deployment, a rephraser conditioned on the rules and the scene image rewrites each incoming instruction once, before the episode begins, and passes the rewrite to the frozen VLA (Fig.~\ref{fig:method}). Only the instruction changes, so the VLA's weights are never modified, and because the rules describe how the policy parses language in general, they transfer to tasks that contributed no evidence.

We first characterize the sensitivity for $\pi_0$ on SIMPLER's Bridge tasks and $\pi_{0.5}$ on LIBERO, through statistically tested single-edit swings and an oracle phrase search that estimates the headroom phrasing alone leaves a frozen policy. Two findings shape the paper. First, oracle phrasing nearly closes the 21-point gap between in-distribution and out-of-distribution tasks, so much of the apparent generalization gap is a phrasing gap. Second, on $\pi_{0.5}$ adversarial phrasing costs 50 points on tasks inside the finetuning set and nothing measurable outside it, which suggests that the policy attends to language only where training has taught it to~\cite{zhou2025liberopro, fei2025liberoplus}.

We evaluate on the rephrase-augmented $\pi_0$ across twelve sealed SIMPLER tasks that contributed no evidence, under adversarial, VLM-generated, human-generated, and canonical phrasings. Rules distilled from eight tasks improve success by 16 to 27\% relative over the no-rephraser baseline on the adversarial and natural sets, with gains concentrated on out-of-distribution tasks. The effect is stable across three applier models and three independent distillations.

In summary, this paper contributes:
\begin{enumerate}
 \item a statistically grounded characterization of language sensitivity in $\pi_0$ and $\pi_{0.5}$ showing that much of the apparent gap between in- and out-of-distribution tasks is attributable to phrasing alone;
 \item a weight-free, once-per-episode method that distills rollout evidence into explicit rephrasing rules which transfer zero-shot to unseen tasks;
 \item an evaluation on held-out tasks across three phrasing styles and three applier models, with a comparison to test-time verification and a replication on $\pi_{0.5}$ and LIBERO.
\end{enumerate}

\section{Related Work}

\textbf{Measuring language sensitivity. }
Embodied red teaming uses a VLM to search for instructions that break language-conditioned robot policies~\cite{karnik2024embodied}. LIBERO-PRO perturbs LIBERO along objects, initial states, instructions, and environments. On the language axis it finds that paraphrased instructions leave accuracy intact, but so do nonsensical ones, because the policy executes the same memorized trajectory regardless~\cite{zhou2025liberopro}. LIBERO-Plus charts the same sensitivity across ten VLAs, where instruction perturbations cost between 14 and 50 points, and reaches the same conclusion that much of the apparent robustness stems from models ignoring the instruction~\cite{fei2025liberoplus}. Our single-edit swings are the complementary, finer-grained observation that when the policy does attend to language, a single word can move success by tens of points, and our oracle search quantifies how much of the apparent generalization gap is in fact attributable to phrasing rather than to genuine task novelty.

\textbf{Model-side remedies. }
The standard remedy is augmentation, which relabels demonstrations with additional phrasings so that the policy sees many surface forms of each
instruction~\cite{kwok2026scaling}, or, as in DIAL and SPRINT, with instructions generated by a VLM or LLM that extend the training vocabulary~\cite{xiao2023dial, zhang2024sprint}. A recent multilingual study reports that training on seven phrasings per task roughly halves phrasing sensitivity~\cite{kirouane2026greek}.
Steerable Policies trains the VLA on synthetic commands at several levels of abstraction so that language can steer low-level behavior~\cite{chen2026steerable}. Stable Language Guidance instead marginalizes decoding over LLM-expanded paraphrases~\cite{zhan2026stable}. All of these change the policy's weights or its decoding, whereas the $\pi_0$ checkpoint we study was already trained with rephrase augmentation~\cite{intact}, and our method targets what remains after such training, without any further training of the policy.

\textbf{Deployment-time remedies. }
Two families leave the weights untouched. \emph{Test-time verification} samples several candidates at every inference step and selects among them with a verifier. RoboMonkey scores candidate actions with a VLM-based verifier~\cite{kwok2025robomonkey}, Do What You Say selects the action sequence whose simulated outcome best matches the VLA's own textual plan~\cite{wu2026dowhatyousay}, and CoVer selects the best phrase-action pair among several rephrasings and actions with a contrastively trained verifier~\cite{kwok2026scaling}. CoVer is our closest comparison, evaluated in \S\ref{sec:cover}, and differs in that we act on phrasing only, never on actions, and rephrase once per episode rather than at every step, so per-step inference cost is unchanged. \emph{Instruction optimization} searches the language input of a frozen policy. SARL learns by online reinforcement
which prompts steer a generalist policy to a given
task~\cite{bhatia2026sarl}, VLA Grounder trains an instruction-to-command
translator against rollout reward for a black-box
VLA~\cite{shodiev2026grounder}, and PALO optimizes language subtask
decompositions for few-shot adaptation~\cite{myers2024palo}. Each optimizes per task through interaction with that task, so what it learns does not transfer. Closest to our method, Jeong et al.\ search for language that improves a frozen VLA and distill the result into a learned language feedback policy that intervenes per state~\cite{jeong2026whattosay}. Our rules are instead explicit and human-readable, distilled once offline from eight tasks, applied once per episode, and applied zero-shot to tasks that contributed no evidence.

\textbf{Language as the interface between VLMs and VLAs. }
RT-2 and PaLM-E showed that web-scale vision-language pretraining
transfers to robot control~\cite{zitkovich2023rt2, driess2023palme}, and
Hi Robot places a VLM above a VLA to turn open-ended prompts into instructions the low-level policy can follow~\cite{shi2025hirobot}, and embodied chain-of-thought trains the VLA itself to reason in language before acting~\cite{zawalski2024ecot}. SayCan
grounds an LLM planner in learned value functions so that proposed plans
respect the robot's skills~\cite{ahn2022saycan}, and Inner Monologue
closes the loop with execution feedback~\cite{huang2022inner}. These systems decide \emph{which} instruction to issue, whereas ours learns which surface forms of an instruction this executor parses reliably and rewrites each incoming instruction into one of those forms. Written guidance as a reusable artifact has precedent in PromptBook, which curates prompting guidance for LLM-written robot code~\cite{arenas2024promptbook}, and DROC, which distills human corrections into a knowledge base retrieved for later tasks~\cite{zha2024droc}. Our rulebook is instead distilled automatically from rollout evidence and targets the phrasing the executor parses.

\begin{figure}
 \vspace*{4pt}%
 \centering
 \includegraphics[width=0.85\columnwidth]{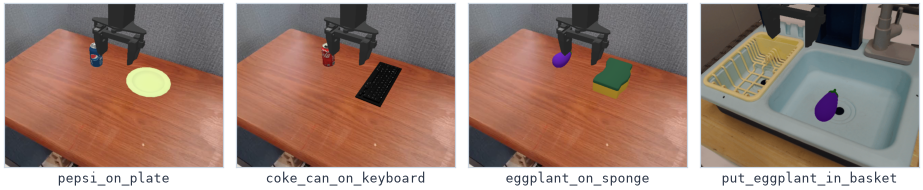}
 \caption{\small Representative SIMPLER Bridge first-frame scenes. Tabletop tasks share one setup and differ only in the source and destination objects; the rightmost panel is the benchmark's sink scene.}
 \label{fig:scenes-bridge}
 \vspace{-0.2cm}
\end{figure}

\begin{figure}
 \vspace*{4pt}%
 \centering
 \includegraphics[width=0.85\columnwidth]{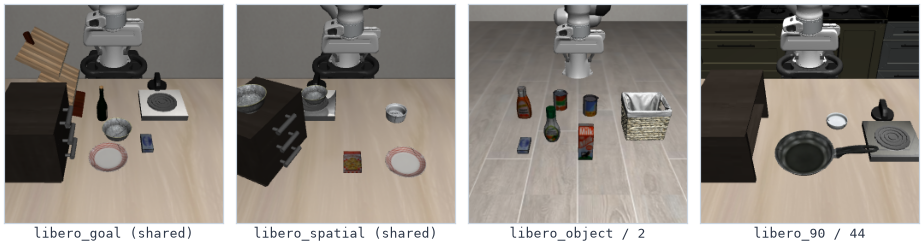}
 \caption{\small Representative LIBERO first-frame scenes. libero\_goal and libero\_spatial share one scene per suite, libero\_object places groceries beside a basket, and libero\_90 scenes vary by task.}
 \label{fig:scenes-libero}
 \vspace{-0.2cm}
\end{figure}

\section{Models and Benchmarks}

We work with two VLAs, $\pi_0$~\cite{pi0} and $\pi_{0.5}$~\cite{pi05}, each with a checkpoint inherited from prior work, namely $\pi_0$ finetuned on Bridge~\cite{intact} and $\pi_{0.5}$ finetuned on LIBERO~\cite{pi05}.  The $\pi_0$ checkpoint was finetuned with rephrase augmentation, that is, on rephrasings of each task, precisely to reduce language sensitivity. It therefore already carries a mitigation, and we study the sensitivity that remains. Success is measured in simulation, at the 60th frame for SIMPLER's Bridge tasks, the benchmark's default horizon, and as described below for LIBERO. Each benchmark provides a set of simulatable tasks, and each task a canonical instruction.

\subsection{Tasks and Canonical Instructions}

\subsubsection{SIMPLER / Bridge}

SIMPLER~\cite{simpler} provides simulated replicas of real-robot evaluation setups. We use its Bridge~\cite{bridge} suite of WidowX tabletop tasks, such as ``put carrot on plate'' and ``put eggplant into yellow basket'' (Fig.~\ref{fig:scenes-bridge}). We use the four Bridge tasks shipped with the benchmark and construct 16 more from benchmark-provided primitives, for 20 simulatable tasks. Eight are \emph{in-vocabulary}, meaning that every noun in the canonical instruction appears in the Bridge training language. Success is determined procedurally by the benchmark. Our \emph{sealed} evaluation set fixes 12 of the 20 ex ante and holds them out of all evidence gathering. Its five in-vocabulary and seven out-of-vocabulary tasks form the in-distribution and out-of-distribution strata reported in our figures.

\subsubsection{LIBERO}

LIBERO~\cite{liu2023libero} consists of 130 simulatable tabletop tasks, such as ``put the moka pot on the stove'', grouped into four suites, shown in Fig.~\ref{fig:scenes-libero}. libero\_spatial, libero\_object, and libero\_goal hold 10 tasks each, and libero\_100 splits into libero\_90 with 90 tasks and libero\_10 with 10. We inherit a $\pi_{0.5}$ finetuned on every suite except libero\_90~\cite{pi05}, so libero\_90 tasks are \emph{out-of-finetune} and the rest \emph{in-finetune}. Success is defined procedurally by the benchmark, and a task counts as successful once any step succeeds before its suite-specific step limit.

\section{Characterizing a VLA's Language Sensitivity}
\label{sec:minimal-pairs}

\subsection{Problem Statement}
\label{sec:problem}

\begin{figure}[!t]
 \vspace*{4pt}%
 \centering
 \includegraphics[width=\columnwidth]{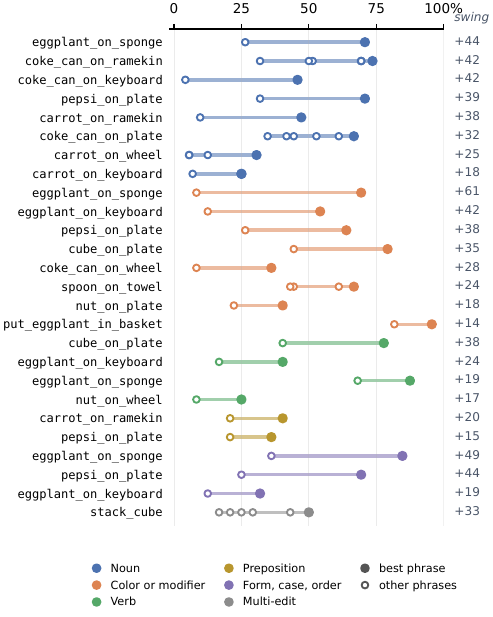}
 \caption{\small Single-edit swings for $\pi_0$ on SIMPLER Bridge: 26 significant sets over 16 tasks, the largest per task and edit category. Each row is one set whose phrases differ by one edit; dots are success rates over n = 72 rollouts (basket task n = 180), the filled dot is the best phrasing, and the right column is the swing in points. Every set has $p < 0.05$ under a two-proportion z test.}
 \label{fig:pairs-bridge}
\end{figure}

\begin{figure}[!t]
 \vspace*{4pt}%
 \centering
 \includegraphics[width=\columnwidth]{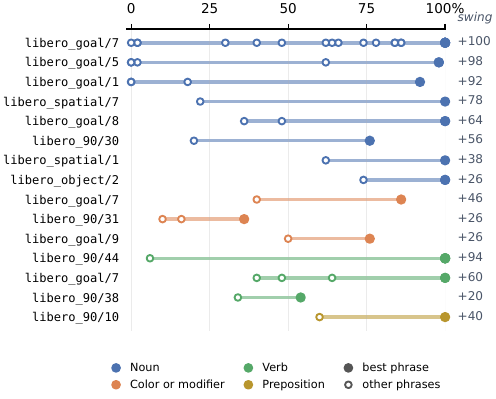}
 \caption{\small Single-edit swings for $\pi_{0.5}$ on LIBERO, formatted as Fig.~\ref{fig:pairs-bridge}: 15 sets over 13 tasks, n = 50 per phrase, with eight further significant sets removed as scene-ambiguous. Swings reach 100\,pp, as ``fire up the stove'' succeeds 100\% of the time and ``turn on the stove'' 6\%.}
 \label{fig:pairs-libero}
\end{figure}

A frozen policy $\pi$ maps an observation and an instruction $\ell$ to actions. For a task $t$, let $S_t(\ell)$ be the success rate of $\pi$ on $t$ under $\ell$ over a fixed set of scene layouts, let $\ell_t$ be the canonical instruction, and let $L_t$ be the admissible rephrasings that preserve the goal. A \emph{single-edit swing} is a pair $\ell, \ell' \in L_t$ that differ by one edit and whose success rates differ significantly, and the \emph{oracle headroom} of $t$ is $\max_{\ell \in L_t} S_t(\ell) - S_t(\ell_t)$, the largest gain available from wording alone. A \emph{rulebook} $R$ defines a rewrite $f_R$ of instructions, distilled from an evidence set $\mathcal{D} = \{(t, \ell, S_t(\ell))\}$ over evidence tasks only. We seek one $R$ that raises $\mathbb{E}_{\ell}[S_t(f_R(\ell))]$ on tasks that contributed no evidence while $\pi$ stays unchanged, $f_R$ is applied once before the first action, and $R$ is fixed before any sealed task is seen. Our tables report the paired difference $S_t(f_R(\ell)) - S_t(\ell)$ over base instructions $\ell$, as defined in \S\ref{sec:stats}. Single-edit swings for $\pi_0$ and $\pi_{0.5}$ are charted in Figs.~\ref{fig:pairs-bridge} and~\ref{fig:pairs-libero}, with the corresponding scenes in Figs.~\ref{fig:scenes-bridge} and~\ref{fig:scenes-libero}.

\subsection{Finding Single-Edit Swings}
\label{sec:swings}

Our search proceeds as follows. We first generate many rephrasings of each task, in fact the same evidence files our rule distillers consume. An LLM is then prompted to propose candidate sets from the evidence, and we evaluate each phrase with many rollouts in simulation. We retain only sets in which the gap in success rate between the best and worst phrase is statistically significant, that is, unlikely to be explained by chance. Each set's phrases and exact p-values are listed in the project repository.\footnote{\url{\artifactsrepo}\label{fn:repo}}
Significance is assessed with a two-proportion z test at $p<0.05$, and for $\pi_{0.5}$ we also require a gap of at least 18 points. No multiple-comparison correction is applied, so with 426 candidate pairs for $\pi_{0.5}$ about 21 significant swings are expected by chance alone. We manually remove sets whose gap could reflect scene ambiguity rather than phrasing, which on $\pi_{0.5}$ removed eight of the 23 significant sets. LIBERO scenes hold multiple objects, so a generalized noun such as ``bottle'' or ``stand'' can match more than one of them, and a set is removed when the ambiguous phrasing is the one that loses, since its deficit may measure scene resolution rather than phrasing. Sets where the ambiguous phrasing wins are kept, and no $\pi_0$ set was removed.
For $\pi_0$ we evaluate each phrase 72 times, on 24 fixed scene layouts with 3 repetitions each, and for $\pi_{0.5}$ once on each of 50 layouts. Common random seeds per layout/repetition pair reduce measurement noise. Sets were harder to find for $\pi_{0.5}$ than for $\pi_0$.

\subsection{Patterns in the Swings}

Several patterns emerge. Some swings hold across tasks and even across models. Replacing plate with dish costs $\pi_0$ 19 points and $\pi_{0.5}$ 56, and ``hot plate'' fails wherever it appears, with 2\% success as a source noun and 0\% as a destination. Other swings do not transfer. Adding ``purple'' to
the eggplant gains 61 points on one task and costs 30 on another, so no
static rule can satisfy both. Swings were also far easier to find on in-finetune tasks. Of the 426 candidate pairs our $\pi_{0.5}$ search evaluated, 44\% of in-finetune pairs showed a significant swing of at least 18 points at $p < 0.05$, versus 5\% of out-of-finetune pairs, which suggests that finetuned performance rests on memorized phrasings that a single word can break. Color shows the clearest asymmetry between the two models. On Bridge, an added color raises success on out-of-vocabulary nouns, as with black keyboard at $+42$, and lowers it on in-vocabulary nouns, as with yellow plate at $-22$. On LIBERO, an added color only ever lowers success, as with black bowl at $-20$. The common thread is plain naming. Generic category words such as can, dish, drink, and thing underperform the specific everyday name, even one the training data never contained, as pepsi exceeds soda by 39 points. Changing Pepsi to pepsi gains 44 points, which suggests sensitivity at the level of tokens rather than words.

\subsection{Oracle Phrase Search}
\label{sec:oracle}

\begin{figure}[!t]
 \vspace*{4pt}%
 \centering
 \includegraphics[width=\columnwidth]{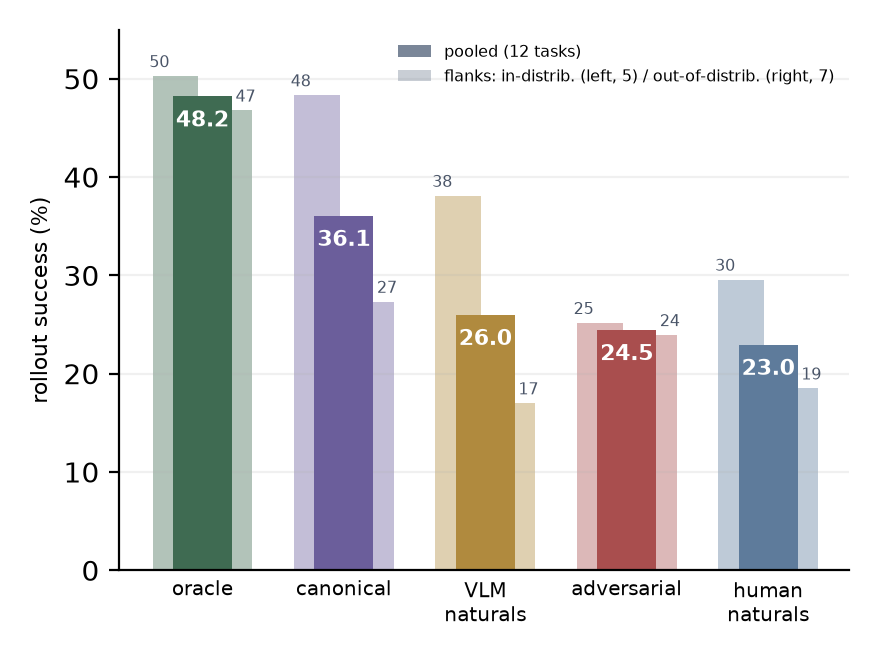}
 \caption{\small Success of the rephrase-augmented $\pi_0$ on the 12 sealed Bridge tasks by phrasing condition, over 24 fixed layouts. Solid bars pool all tasks; thin flanks are the in-distribution (left, 5 tasks) and out-of-distribution (right, 7) strata. Repetitions per layout: oracle and canonical 12, adversarial 2, naturals 1. Oracle phrasing shrinks the 21-point strata gap to 3 points.}
 \label{fig:pi0-conditions}
\end{figure}

\begin{figure}[!t]
 \vspace*{4pt}%
 \centering
 \includegraphics[width=\columnwidth]{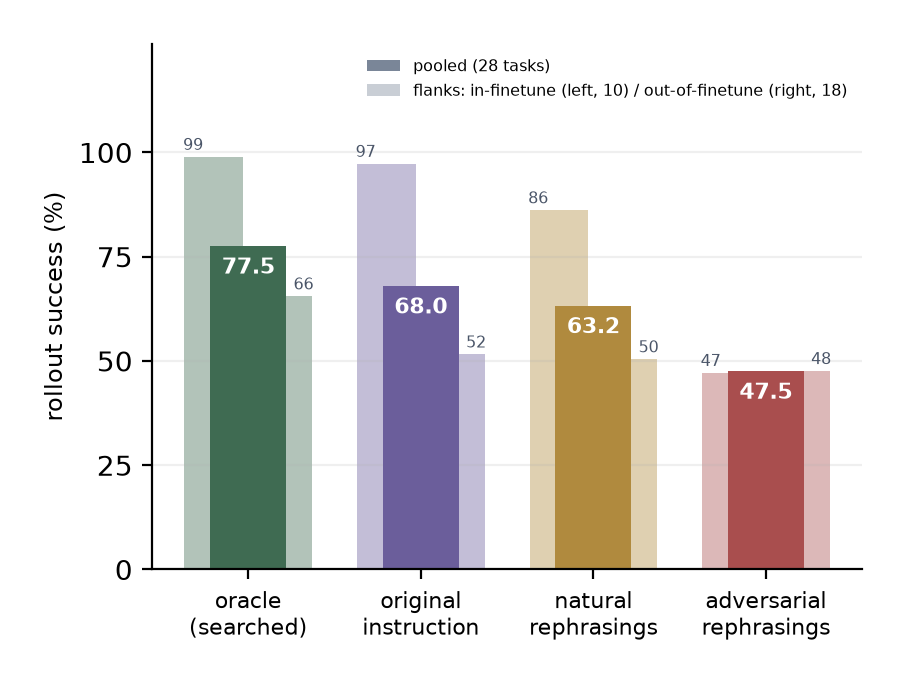}
 \caption{\small Success of $\pi_{0.5}$ on LIBERO by phrasing condition over 50 layouts per phrase. Solid bars pool all 28 tasks; thin flanks are the in-finetune (left, 10 tasks) and out-of-finetune (right, 18) halves. In-finetune, adversarial rephrasing costs 50\,pp (97 to 47); out-of-finetune, the canonical, natural, and adversarial tiers are indistinguishable (52 / 50 / 48). Oracle search finds headroom in both regimes (99 / 66).}
 \label{fig:pi05-fourtier}
\end{figure}

To quantify sensitivity further, we compare each task's best-performing phrase with its canonical instruction. In each of three rounds, a VLM generates 16 rephrasings of the task, each is scored by several rollouts, and the next round conditions on all scored phrases so far. We then select the best-scoring phrase.
Rephrase candidates are generated by gemini-pro-latest. For $\pi_0$, each candidate is measured on layouts 0--17 with $n = 36$, and selection uses these layouts only. The selected phrase is then re-measured 12 times per layout on all 24 layouts, which yields the oracle bars in Fig.~\ref{fig:pi0-conditions}. The held-out layouts 18--23 alone, with $n = 72$, give the same headroom, so selection does not inflate it. For $\pi_{0.5}$, candidates are screened on 5--10 layouts, and the oracle bar in Fig.~\ref{fig:pi05-fourtier} pools each winner's re-measurement on disjoint layouts with $n = 10$--$20$ per task.
In Figs.~\ref{fig:pi0-conditions} and~\ref{fig:pi05-fourtier}, oracle phrases reach significantly higher success rates than the canonical, adversarial, and natural sets for both models. By this measure $\pi_0$ is more sensitive to phrasing than $\pi_{0.5}$, consistent with the swing search of \S\ref{sec:swings}.

\section{Reducing Language Sensitivity with LLM-Distilled Rephrasing Rules}

The method, summarized in Fig.~\ref{fig:method}, proceeds in three stages. \emph{Evidence gathering} scores many rephrasings of the evidence tasks on the frozen VLA, \emph{rule distillation} turns the scored phrasings into a rulebook $R$ once and offline, and \emph{rule application} rewrites each incoming instruction under $R$ once at deployment. Unlike per-task optimization, the distilled rules must hold across all tasks at once.

\subsection{Evidence Gathering}
\label{sec:evidence}

Rules are derived from the \emph{evidence tasks}, which comprise the eight simulatable Bridge tasks outside the sealed set and non-simulatable tasks with real-world expert trajectories. For each evidence task, a VLM, gemini-pro-latest, generates seven \emph{natural} rephrasings, worded the way a person might request the task, and two \emph{adversarial} rephrasings, which are convoluted and overly complicated. Fig.~\ref{fig:method} shows one of each for ``put carrot on plate''. Every phrase measured during oracle search is also included, and most of these are benign. The prompt used for generation is provided in the project repository.\footref{fn:repo}
The non-simulatable tasks are the 208 remaining tasks of a 228-task pool built from BridgeData V2's language-annotated expert demonstrations~\cite{bridge}. The evidence $\mathcal{D}$ therefore holds 5{,}453 scored task--phrase pairs drawn from these 228 evidence tasks.

\subsubsection{Scoring Phrases}

Rule distillation requires a score per phrase so that phrases are comparable. For simulatable tasks, each task/phrase pair is rolled out 36 times, on 18 layouts with 2 repetitions each, and scored by its success rate. For non-simulatable tasks the score is a proxy, the gripper error between the VLA's predicted action and the ground-truth action, averaged over 4 frames in each of 16 expert trajectories, for 64 measurements per phrase. On simulatable tasks this proxy correlates reasonably well with rollout success.
The error is the mean absolute difference on the gripper dimension of the predicted action chunk, in raw action units, averaged over the chunk horizon and over four decoding draws. Across 434 task--phrase pairs over 15 task variants, the proxy correlates with rollout success at Pearson $r = 0.54$ within task.

\subsection{Rule Distillation}

An LLM, Claude, derives the rules from the evidence with the prompt provided in the project repository.\footref{fn:repo} On Bridge, three rulebook types are distinguished by the evidence they see, namely \emph{in-finetune}, \emph{out-of-finetune}, and \emph{both}. Because distillation is stochastic, each type is distilled three times and the results are averaged. Rulebooks contain 10 to 20 rules, ranging from general constraints such as \textit{``never swap a common noun for its sibling: keep `plate', not `dish' ''} to lexical substitutions such as \textit{```onto' becomes `on', `into' becomes `in' ''} for the two prepositions.

For rulebooks that see in-finetune phrases, the distiller also receives the full set of finetuning instructions so that it can use corpus statistics such as object frequencies and common phrase structures. The first rulebook of each type was distilled with Claude Fable, model claude-fable-5, on 2026-09-03, and the remaining two with Claude Opus 5, model claude-opus-5, on 2026-09-05. The results do not indicate that this choice affected performance.
Each distillation is a multi-agent panel of three lens distillers, an adversarial critic, and a synthesizer, run once without iteration. The replicate panels permute the lens order and see the same evidence files. Requests used the API-default temperature of 1.0 and the default adaptive thinking depth.
All nine rulebooks are available in full in the project repository for inspection and reuse.\footref{fn:repo}

\subsection{Rule Application}

At deployment, the rewrite $f_R$ is applied once per episode, before the robot moves, by one of three \emph{appliers}, namely Claude, Gemini, or the open-weights Qwen3.5-9B~\cite{qwen}. Besides the instruction and the rulebook $R$, the applier receives a reasoning trace produced by Gemini 3.5 Flash at temperature 0.4 from the incoming phrase and the scene image. The applier itself does not receive the image. The appliers are Claude Fable 5 at effort high, or max on the human-generated set, Gemini through the gemini-pro-latest alias as accessed in September 2026 with a 16{,}384-token thinking budget, and Qwen3.5-9B with greedy decoding. The applier applies the rules greedily, and its output goes directly to the unchanged VLA.
Before the robot moves, the pipeline makes two model calls per instruction. Trace generation takes about 2.5\,s, and rule application about 10\,s for Claude, about 17\,s for Gemini, and 1--2\,s for the local Qwen. The rewrite is computed once and reused across all layouts and repetitions.

\section{Evaluating the Rulebooks}
\label{sec:rules-results}

We evaluate rulebooks on the rephrase-augmented $\pi_0$ over the 12 sealed Bridge tasks and 24 fixed layouts, and \S\ref{sec:libero-repl} replicates the study on $\pi_{0.5}$ and LIBERO. The layouts are SIMPLER's deterministic object-initialization episodes 0--23, shared by every arm. The sealed split was fixed weeks before any rulebook evidence was generated, and only non-sealed tasks ever entered the evidence bank used for distillation.

\subsection{Phrase Sets}
\label{sec:phrase-sets}

We evaluate on four phrase sets: adversarials, VLM-generated naturals, human-generated naturals, and canonicals. The adversarial and natural sets follow the two styles of \S\ref{sec:evidence}, here produced by Gemini 3.5 Flash from the scene image, with the prompt in the project repository. Human-generated naturals were collected through a survey of acquaintances of the authors, who watched a successful video demonstration of each task and were asked to request it of an 8-year-old, an adult, and a robot, yielding 78, 159, and 155 phrases respectively.
We received 37 submissions from respondents aged 11 to 85, of mixed gender, and with AI familiarity ranging from none to professional. Respondents never saw the canonical instruction, and every field was optional. Phrases are kept verbatim, including typographical errors. Cleaning dropped only eight placeholder or incomplete entries and collapsed double-posted duplicates, leaving 392 phrases, 363 of them unique.
Per task we evaluate 6 adversarial phrases, roughly 16 VLM-generated naturals, and roughly 30 human-generated naturals, for 72, 186, and 363 phrases in total. We also evaluate each task's canonical instruction, but with one phrase per task the canonical results in Fig.~\ref{fig:dashboard} are statistically underpowered and should be read with caution.

\subsection{Baselines}

We compare against three baselines. The \emph{no-rephraser} baseline passes the instruction to the VLA unchanged. The \emph{no-rules rephraser} runs the same applier with a single generic rule (see the project repository), isolating the effect of the distilled rules from rephrasing itself. CoVer~\cite{kwok2026scaling} uses a contrastively trained verifier to select, at every step, the highest-scoring phrase/action pair from 40 candidates (8 rephrasings $\times$ 5 actions per step).

\subsection{Statistical Tests}
\label{sec:stats}

For a comparison between two arms, the paired unit is the \emph{base instruction}. For base $i$, $d_i = c_i - b_i$ is the difference between the two arms' success rates over the same fixed grid of scene layouts, which for a rulebook arm against the no-rephraser is $S_t(f_R(\ell_i)) - S_t(\ell_i)$ in the notation of \S\ref{sec:problem}. Rulebook arms average their three draws before pairing. Common random numbers make the two arms face identical
scenes, so scene difficulty cancels within each $d_i$. Success is \emph{base-weighted} throughout, including in Fig.~\ref{fig:dashboard}. Rollouts are averaged per task and phrase and then over base instructions, so each base counts equally however many bases an applier collapses onto the same rewrite. The reported
p-value is a two-sided sign-flip permutation test on the mean of the $d_i$, with 20{,}000 resamples or exact enumeration when $n \le 12$. The number of bases is the sample size, namely $n = 186$ for VLM-generated naturals, $363$ for human-generated naturals, $72$ for adversarials, and $12$ for canonicals. Layouts and repetitions sharpen each $d_i$ but do not add to $n$. The LIBERO replication uses the same paired construction with $n = 200$ bases and 10{,}000 permutations. Weighting
duplicate human submissions by multiplicity shifts pooled rates by less
than a point and changes no conclusion.
Tables~\ref{tab:pscores-baseline} and~\ref{tab:pscores-scaffold} report each effect as a relative change, the mean paired difference as a percent of the baseline's pooled rate. For the no-rephraser these rates are 24.5\% on adversarials, 23.0\% on human-generated naturals, and 26.0\% on VLM-generated naturals.

\subsection{Results}

\begin{figure*}[!t]
 \vspace*{4pt}%
 \centering
 \includegraphics[width=0.9\textwidth]{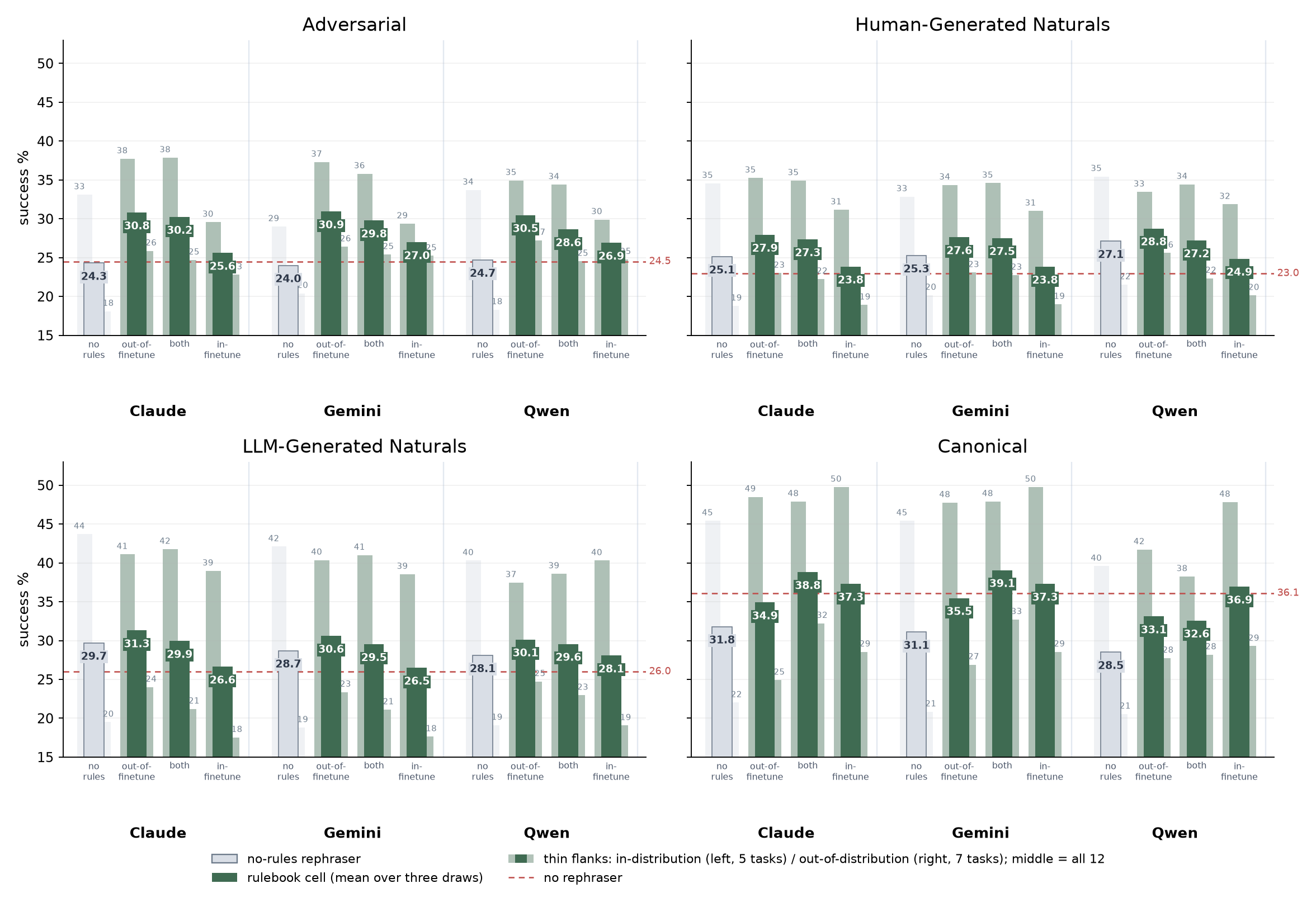}
 \caption{\small Main results: success of the rephrase-augmented $\pi_0$ on the 12 sealed SIMPLER tasks, one panel per phrase set and one bar group per applier. Dashed red = no-rephraser baseline; grey = no-rules rephraser; greens = the out-of-finetune, both, and in-finetune rulebooks, each the mean over three draws; thin flanks = in-distribution (left, 5 tasks) and out-of-distribution (right, 7) strata. Out-of-finetune rulebooks are strongest on every non-canonical set, with gains concentrated out-of-distribution; the canonical panel, with one phrase per task, is statistically underpowered and should be read with caution.}
 \label{fig:dashboard}
\end{figure*}

The out-of-finetune rulebook is consistently the best type. The one exception is the canonical set, where the in-finetune rulebook is best, but that set is underpowered. Gains concentrate on out-of-distribution tasks. With out-of-finetune rulebooks, the Claude applier improves on the no-rephraser baseline by 20 to 26\% relative across the adversarial and natural sets, as Table~\ref{tab:pscores-baseline} shows, and on the no-rules rephraser by 26\% on adversarials, 11\% on human-generated naturals, and 4.7\% on VLM-generated naturals, as Table~\ref{tab:pscores-scaffold} shows. Fig.~\ref{fig:dashboard} gives the full results. Across the three draws of a rulebook type, pooled success varies by 0.7 to 5.1 points, with a median range of 3.0, small relative to these effects.

\begin{table}[!t]
 \caption{\small Relative improvement of the out-of-finetune rulebook over the no-rephraser baseline: mean paired difference as a percent of the baseline rate, with permutation-test p-values (\S\ref{sec:stats}); n is the number of base instructions.}
 \label{tab:pscores-baseline}
 \centering
 \footnotesize
 \setlength{\tabcolsep}{4pt}
 \begin{tabular}{@{}lccc@{}}
\toprule
 & Adversarial & Human-Generated & VLM-Generated \\
 & (n=72) & Naturals (n=363) & Naturals (n=186) \\
\midrule
Claude & +26.0\%~(p=0.0083) & +21.7\%~(p<0.0001) & +20.5\%~(p<0.0001) \\
Gemini & +26.5\%~(p=0.0074) & +20.3\%~(p<0.0001) & +17.8\%~(p<0.0001) \\
Qwen & +24.4\%~(p=0.0112) & +25.2\%~(p<0.0001) & +16.1\%~(p=0.0001) \\
\bottomrule
\end{tabular}

\end{table}

\begin{table}[!t]
 \caption{\small As Table~\ref{tab:pscores-baseline}, but against the no-rules rephraser, isolating the effect of the distilled rules from rephrasing alone.}
 \label{tab:pscores-scaffold}
 \centering
 \footnotesize
 \setlength{\tabcolsep}{4pt}
 \begin{tabular}{@{}lccc@{}}
\toprule
 & Adversarial & Human-Generated & VLM-Generated \\
 & (n=72) & Naturals (n=363) & Naturals (n=186) \\
\midrule
Claude & +26.6\%~(p=0.0001) & +11.2\%~(p<0.0001) & +4.7\%~(p=0.0924) \\
Gemini & +29.1\%~(p=0.0001) & +9.3\%~(p=0.0001) & +6.7\%~(p=0.0167) \\
Qwen & +23.1\%~(p=0.0001) & +6.0\%~(p=0.0201) & +7.3\%~(p=0.0271) \\
\bottomrule
\end{tabular}

\end{table}

\subsection{Comparison with Test-Time Verification}
\label{sec:cover}

Table~\ref{tab:cover} compares CoVer with our baselines. All arms use identical Gemini reasoning traces, and the no-rules rephraser column averages the three appliers. Running CoVer's framework with the verifier disabled reproduces its published baseline results. CoVer's native metric is first success within an episode, whereas we report final-state success. Auditing our results under first success improved all arms by similar amounts and changed no conclusion.

\begin{table}[!t]
 \caption{\small CoVer against our two baselines on the sealed tasks: pooled success (\%) at step 60, all arms using identical Gemini reasoning traces. CoVer trails both baselines on every set.}
 \label{tab:cover}
 \centering
 \footnotesize
 \setlength{\tabcolsep}{5pt}
 \begin{tabular}{@{}lccc@{}}
\toprule
 & no rephraser & no-rules rephraser & CoVer \\
\midrule
Adversarial & 24.5 & 24.3 & 21.6 \\
VLM-Generated Naturals & 26.0 & 28.8 & 22.8 \\
Canonical & 36.1 & 30.5 & 26.4 \\
\bottomrule
\end{tabular}

\end{table}

\subsection{Replication on $\pi_{0.5}$ and LIBERO}
\label{sec:libero-repl}

We repeated the pipeline on $\pi_{0.5}$ with a preregistered sealed set of 20 tasks, 10 in-finetune and 10 out-of-finetune, 10 VLM-generated natural rephrasings per task, evidence from the non-sealed single-edit pairs only, and the Gemini applier, for $n = 200$ bases over 50 layouts per arm. Three independently distilled rulebooks lift in-finetune success from 93.6\% to 97.8\% on average, with $p \le 0.005$, and leave out-of-finetune success unchanged, with $p > 0.4$. The no-rules rephraser gains nothing in-finetune but 1.6 points out-of-finetune, with $p = 0.006$. The rulebooks' gains are the repairs that the evidence licenses, measured as per-phrase gains averaged across rulebooks. Rewriting hot plate, hotplate, or burner to stove gains 45 points on the roughly 9 of the 200 test phrases that used them, and rewriting dish to plate gains 24 points. Smaller but more broadly applied, rewriting onto to on and into to in gains 2.5 points over roughly 50 phrases, and two of the three rulebooks also rewrite grab or lift to pick up, gaining 2.4 points over roughly 15 phrases. A prompt that required each rule to hold across all tasks produced a rulebook that changed little and gained nothing, whereas asking for rules that help on average produced the rulebooks above, two distilled by Claude Fable 5.1 and one by Claude Opus 5.

\section{Conclusion}
\label{sec:conclusion}

Several directions remain open for future work. Rulebooks distilled from more tasks may yield larger gains, and a more formal reinforcement-learning treatment of phrasing is appealing but would demand far more simulatable tasks. Our in-finetune rulebooks underperformed, possibly because of a weak proxy score, whereas the LIBERO rulebooks, whose evidence was entirely rollout-scored, did not.  We evaluated the method on two VLAs and only in simulation, so real robots are a natural next step. Most importantly, language sensitivity should be measured systematically across models. Explaining it, by attributing the sensitivity to a cause, would be the most valuable outcome of this line of work.

\section*{ACKNOWLEDGMENT}

We thank Amber Kleiner for designing Fig.~\ref{fig:method}.
The code was written with Claude (Anthropic) under the authors'
direction and review. The paper text was written by the authors with assistance from
Claude: it helped compile the Related Work section, drafted the figure
and table captions and the Statistical Tests subsection.
The authors
reviewed and are responsible for all content.

\balance
\bibliographystyle{IEEEtran}
\bibliography{references}

\end{document}